\documentclass[11pt]{article}
\usepackage{acl2016}
\usepackage{times}
\usepackage{latexsym}
\usepackage{url}
\usepackage{bm}
\usepackage{tabularx}
\usepackage{ragged2e}
\usepackage{amsmath}
\newcolumntype{Y}{>{\RaggedRight\arraybackslash}X}
\usepackage{booktabs}

\usepackage{graphicx}
\usepackage{subcaption}
\usepackage{amssymb}
\usepackage[titletoc,title]{appendix}
\usepackage[acronym]{glossaries}

\usepackage[colorinlistoftodos,prependcaption,textsize=tiny]{todonotes}

\newacronym{MemNN}{MemNN}{Memory Network}
\newacronym{ptrnet}{Ptr-Net}{Pointer Network}
\newacronym{psr}{AS Reader}{Attention Sum Reader}

\newcommand{\attvis}[2]{\definecolor{att}{rgb}{1, #2, #2} \colorbox{att}{#1}}

\newcommand{\MARTIN}[1]{{\color{black}#1}} 
\newcommand{\RUDA}[1]{{\color{black}#1}} 
\newcommand{\ONDRA}[1]{{\color{black}#1}} 

\newcommand{\RUDAA}[1]{{\color{black}#1}} 

\makeglossaries

\newglossaryentry{MenNN}
{
  name=MemNN,
  description={is a programmable machine that receives input,
               stores and manipulates data, and provides
               output in a useful format}
}

\title{Text Understanding with the Attention Sum Reader Network}
\author{Rudolf Kadlec, Martin Schmid, Ondrej Bajgar \& Jan Kleindienst \\
IBM Watson\\
V Parku 4, Prague, Czech Republic\\
\texttt{\{rudolf\_kadlec,martin.schmid,obajgar,jankle\}@cz.ibm.com} \\ 
}

\newcommand{\fwd}[1]{\overrightarrow{#1}}
\newcommand{\back}[1]{\overleftarrow{#1}}

\newcommand{\ctxTransducer}{f}
\newcommand{\queryEncoder}{g}

\newcommand{\embed}{e}

\newcommand{\querySeq}{\mathbf{q}}

\newcommand{\documentSeq}{\mathbf{d}}

\newcommand{\answer}{a}
\newcommand{\vocabMat}{\mathbf{V}}

\newcommand{\dg}{\textsuperscript{\dag}}
\newcommand{\ddg}{\textsuperscript{\ddag}}
\newcommand{\dddg}{$^\sharp$}

\aclfinalcopy % Uncomment for camera-ready version
\def\aclpaperid{413} %  Enter the acl Paper ID here

\begin{document}

\maketitle

\begin{abstract}
\MARTIN
{
%Two large-scale cloze-style context-question-answer datasets have been introduced recently: i) the CNN and Daily Mail news data and ii) the Children's Book Test. Thanks to the size of these datasets, the associated task is well suited for deep-learning techniques that seem to outperform all alternative approaches. We present a new simple attention-based architecture that is tailor made for such question-answering problems. While our model is simpler than models previously proposed for these tasks, it outperforms them by a large margin.

Several large cloze-style context-question-answer datasets have been introduced recently: the CNN and Daily Mail news data and the Children's Book Test.
Thanks to the size of these datasets, the associated text comprehension task is well suited for deep-learning techniques that currently seem to outperform all alternative approaches.
We present a new, simple model that uses attention to directly pick the answer from the context as opposed to computing the answer using a blended representation of words in the document as is usual in similar models.
This makes the model particularly suitable for question-answering problems where the answer is a single word from the document.}
\RUDAA{Ensemble of our models sets new state of the art on all evaluated datasets.}

%\todo[inline]{Our model outperforms models previously proposed for these tasks by a large margin.}

%Moreover, our single model accuracy is better than even the best previously reported ensembles.
%Fusing multiple instances of our model leads to further improvements in accuracy.

\end{abstract}

\section{Introduction}

Most of the information humanity has gathered up to this point is stored in the form of plain text. 
Hence the task of teaching machines how to understand this data is of utmost importance in the field of Artificial Intelligence.
One way of testing the level of text understanding is simply to ask the system questions for which the answer can be inferred from the text.
A well-known example of a system that could make use of a huge collection of unstructured documents to answer questions is for instance 
IBM's Watson system used for the Jeopardy challenge~\cite{Ferrucci2010}.

\begin{figure}[t!]
\small
\centering
\begin{tabular}{| p{\dimexpr0.47\textwidth-2\tabcolsep-\arrayrulewidth\relax}|   }
\hline
\\[-5pt]
\textbf{Document:} What was supposed to be a fantasy sports car ride at Walt Disney World Speedway turned deadly when a Lamborghini crashed into a guardrail.
The crash took place Sunday at the Exotic Driving Experience, which bills itself as a chance to drive your dream car on a racetrack.
The Lamborghini's passenger, 36-year-old Gary Terry of Davenport, Florida, died at the scene, Florida Highway Patrol said.
The driver of the Lamborghini, 24-year-old Tavon Watson of Kissimmee, Florida, lost control of the vehicle, the Highway Patrol said. (...)
\\[5pt]
\hline
%\begin{figure}[h!]

% \caption{See that the ....}
% \label{fig:attention_1}
%\end{figure}
\\[-5pt]
\textbf{Question:} Officials say the driver, 24-year-old Tavon Watson, lost control of a \_\_\_\_\_\_\_ 
\\[5pt]
\hline
\\[-5pt]
\textbf{Answer candidates}: Tavon Watson, Walt Disney World Speedway, Highway Patrol, Lamborghini, Florida, (...)
\\[5pt]
\hline
\\[-5pt]
\textbf{Answer:} Lamborghini
\\[5pt]
\hline
\end{tabular}
\caption{Each example consists of a context document, question, answer cadidates and, in the training data, the correct answer. This example was taken from the CNN dataset~\protect\cite{hermann2015teaching}. Anonymization of this example that makes the task harder is shown in Figure~\ref{tab:goodEx}.}

\label{tab:example_question}
\end{figure}

\begin{table*}[t]
\begin{center}
   \resizebox{\textwidth}{!}{
%  \centering
  \begin{tabular}[t]{@{}l@{~}r@{~~}r@{~~}r@{}l@{}r@{~~}r@{~~}r@{}l@{}r@{~~}r@{~~}r@{}l@{}r@{~~}r@{~~}r}
    \toprule
    & \multicolumn{3}{c}{{\bf CNN}} &\phantom{aa}& \multicolumn{3}{c}{{\bf
Daily Mail}} &\phantom{aa}& \multicolumn{3}{c}{{\bf
CBT CN}}&\phantom{aa}& \multicolumn{3}{c}{{\bf CBT NE}}\\ %Common Nouns}}&\phantom{aa}& \multicolumn{3}{c}{{\bf CBT Named Entities}}\\
    \cmidrule{2-4} \cmidrule{6-8} \cmidrule{10-12} \cmidrule{14-16}
    & train & valid & test && train & valid & test && train & valid & test&& train & valid & test \\
    \midrule
    \# queries   & 380,298  & 3,924 & 3,198 && 879,450 & 64,835 & 53,182  && 120,769 & 2,000 & 2,500 && 108,719 & 2,000 & 2,500\\
    Max \# options & 527   & 187   & 396   && 371     & 232    & 245 && 10     & 10    & 10 && 10     & 10    & 10\\
    Avg \# options & 26.4  & 26.5  & 24.5  && 26.5    & 25.5   & 26.0 && 10     & 10    & 10&& 10     & 10    & 10\\
    Avg \# tokens  & 762    & 763   & 716   && 813     & 774    & 780  && 470     & 448    & 461 && 433     & 412    & 424\\
    Vocab. size & \multicolumn{3}{c}{{118,497}} && \multicolumn{3}{c}{{208,045}} && \multicolumn{3}{c}{{53,185}}&& \multicolumn{3}{c}{{53,063}}\\
    \bottomrule
  \end{tabular}
  }
\end{center}  
  
  \caption{Statistics on the 4 data sets used to evaluate the model. CBT CN stands for CBT Common Nouns and CBT NE stands for CBT Named Entites. Statistics were taken from \protect\cite{hermann2015teaching} and the statistics provided with the CBT data set.}
    \label{tab:corpus-stats}
\end{table*}

Cloze-style questions~\cite{taylor1953cloze}, i.e. questions formed by removing a phrase from a sentence, are an appealing form of such questions (for example see Figure~\ref{tab:example_question}).
While the task is easy to evaluate, one can vary the context, the question sentence or the specific phrase missing in the question to dramatically change the task structure and difficulty.

One way of altering the task difficulty is to vary the word type being replaced, as in~\cite{hill2015goldilocks}.
The complexity of such variation comes from the fact that the level of context understanding needed in order to correctly predict different types of words varies greatly.
While predicting prepositions can easily be done using relatively simple models with very little context knowledge, predicting named entities requires a deeper understanding of the context.

Also, as opposed to selecting a random sentence from a text as in~\cite{hill2015goldilocks}), the question can be formed from a specific part of the document, such as a short summary or a list of tags. Since such sentences often paraphrase in a condensed form what was said in the text, they are particularly suitable for testing text comprehension~\cite{hermann2015teaching}.

An important property of cloze-style questions is that a large amount of such questions can be automatically generated from real world documents. This opens the task to data-hungry techniques such as deep learning. 
This is an advantage compared to smaller machine understanding datasets like MCTest~\cite{Richardson2013} that have only hundreds of training examples and therefore the best performing systems usually rely on hand-crafted features~\cite{Sachan2015,Narasimhan2015a}.

In the first part of this article we introduce the task at hand and the main aspects of the relevant datasets. 
Then we present our own model to tackle the problem. Subsequently we compare the model to previously proposed architectures and finally describe the experimental results on the performance of our model.
%The rest of the paper is structured as follows.
%1. Formalizaton of the problem  
%1. Summary of the corresponding datasets 
%3. Motivation and formal description of our model
%4. Related work including discussion of similarities and differences
%5. Experimental results

\section{Task and datasets}

In this section we introduce the task that we are seeking to solve and relevant large-scale datasets that have recently been introduced for this task.

\subsection{Formal Task Description}
The task consists of answering a cloze-style question, the answer to which depends on the understanding of a context document provided with the question. The model is also provided with a set of possible answers from which the correct one is to be selected. This can be formalized as follows:

The training data consist of tuples $( \querySeq, \documentSeq, \answer, A )$,
where $\querySeq$ is a question, $\documentSeq$ is a document that contains the answer to question $\querySeq$, $A$ is a set of possible answers and $a \in A$ is the ground truth answer. 
Both $\querySeq$ and $\documentSeq$ are sequences of words from vocabulary $V$. We also assume that all possible answers are words from the vocabulary, that is $A \subseteq V$, and that the ground truth answer $a$ appears in the document, that is $\answer \in \documentSeq$.
% Recently, there have been introduced new large-scale, context-question-answer dataset~\cite{hermann2015teaching,hill2015goldilocks}.
% Due to nature of the construction, the questions are cloze-style questions.

\subsection{Datasets}

%Three large datasets have recently been introduced for training and evaluating models built for the task described above. 
We will now briefly summarize important features of the datasets.

\subsubsection{News Articles --- CNN and Daily Mail}
The first two datasets\footnote{The CNN and Daily Mail datasets are available at \url{https://github.com/deepmind/rc-data}}~\cite{hermann2015teaching} were constructed from a large number of news articles from the CNN and Daily Mail websites.
The main body of each article forms a context, while the cloze-style question is formed from one of short highlight sentences, appearing at the top of each article page.
Specifically, the question is created by replacing a named entity from the summary sentence (\emph{e.g. ``Producer X will not press charges against Jeremy Clarkson, his lawyer says.''}).

Furthermore the named entities in the whole dataset were replaced by anonymous tokens which were further shuffled for each example so that the model cannot build up any world knowledge about the entities and hence has to genuinely rely on the context document to search for an answer to the question.
%and the resulting size of the datasets is $400,000$ and $900,000$ tuples respectively \footnote{TODO available here}.

\RUDAA{
Qualitative analysis of reasoning patterns needed to answer questions in the CNN dataset together with human performance on this task are provided in~\cite{chen2016thorough}.
}

\subsubsection{Children's Book Test}
The third dataset\footnote{The CBT dataset is available at \url{http://www.thespermwhale.com/jaseweston/babi/CBTest.tgz}}, the Children's Book Test (CBT)~\cite{hill2015goldilocks}, is built from books that are freely available 
thanks to Project Gutenberg\footnote{\url{https://www.gutenberg.org/}}.
Each context document is formed by $20$ consecutive sentences taken from a children's book story.
Due to the lack of summary, the cloze-style question is then constructed from the subsequent ($21$\textsuperscript{st}) sentence.

%By varying the type of replaced entity, 
One can also see how the task complexity varies with the type of the omitted word (named entity, common noun, verb, preposition). 
\cite{hill2015goldilocks} have shown that while standard LSTM language models have human level performance on predicting verbs and prepositions,
they lack behind on named entities and common nouns. In this article we therefore focus only on predicting the first two word types. %named entities and common nouns, since the other two word types are almost
%The so far best reported architecture on these word types was derived from Memory Networks~\cite{Sukhbaatar2015} extended with a so called \textit{self supervision} mechanism.

Basic statistics about the CNN, Daily Mail and CBT datasets are summarized in Table~\ref{tab:corpus-stats}.

\vspace{10pt} % Forcing the title to jump to the next column

\section{Our Model --- Attention Sum Reader}

\begin{figure*}[ht]
  \centering
  \includegraphics[width=5in]{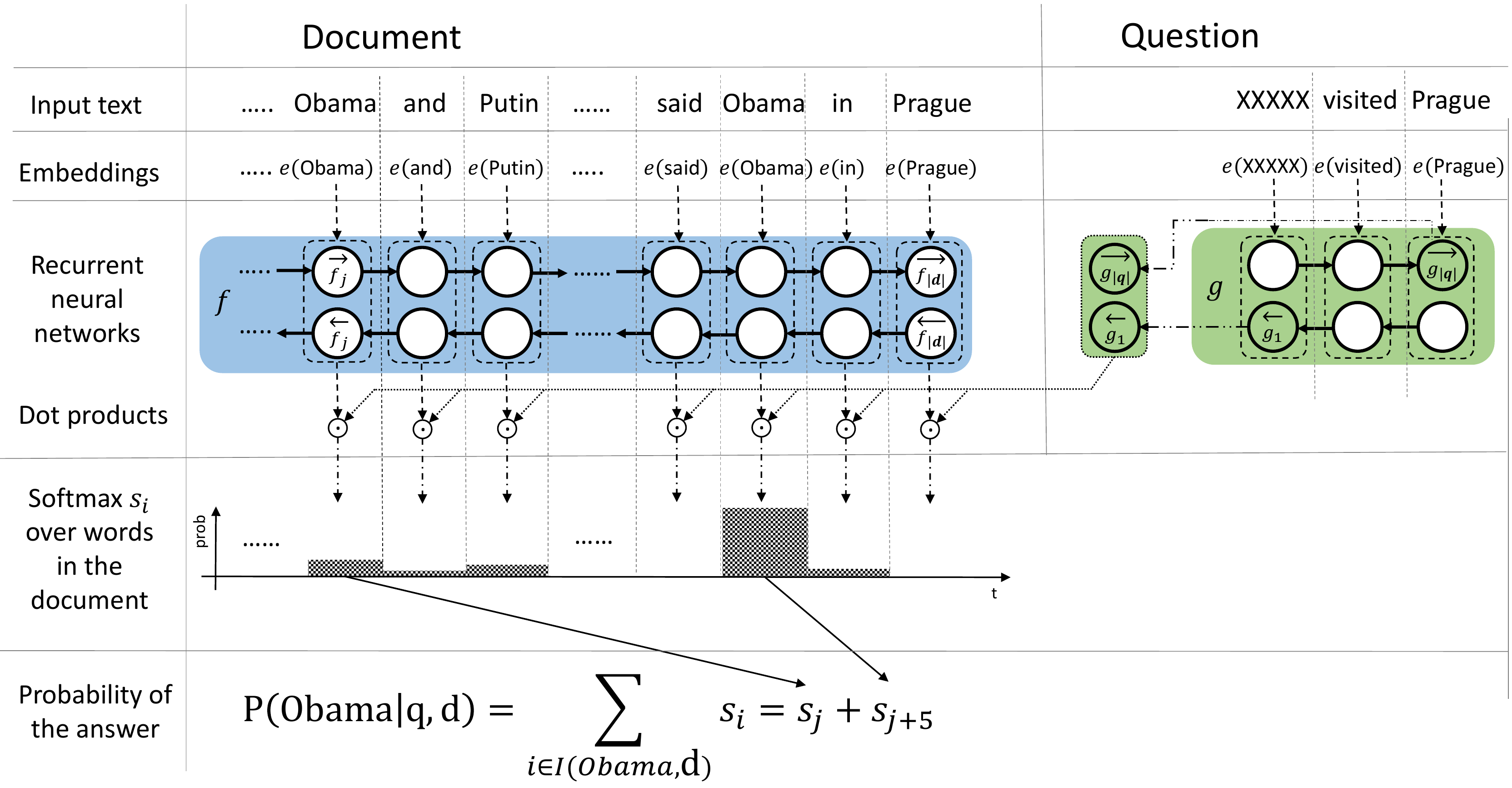}
  \caption   
  {
  Structure of the model.
  }
  \label{fig:model-structure}
\end{figure*}

\MARTIN
{

%\subsection{Motivation}
Our model called the \emph{\gls{psr}}\footnote{\RUDAA{Our implementation of the \gls{psr} is available at \url{https://github.com/rkadlec/asreader}}} is tailor-made to leverage the fact that the answer is a word from the context document.
This is a double-edged sword.
While it achieves state-of-the-art results on all of the mentioned datasets (where this assumption holds true), it cannot produce an answer which is not contained in the document.
Intuitively, our model is structured as follows:
\begin{enumerate}
\item We compute a vector embedding of the query.
\item We compute a vector embedding of each individual word in the context of the whole document (\emph{contextual embedding}).
\item Using a dot product between the question embedding 
and the contextual embedding of each occurrence of a candidate answer in the document, we select the most likely answer.
\end{enumerate}
}

\subsection{Formal Description}

Our model uses one word embedding function and two encoder functions. The word embedding function $e$ translates words into vector representations.
The first encoder function is a document encoder $\ctxTransducer$ that encodes every word from the document $\documentSeq$ in the context of the whole document. We call this the  \emph{contextual embedding}. For convenience we will denote the contextual embedding of the $i$-th word in $\documentSeq$ as $\ctxTransducer_i(\documentSeq)$. 
The second encoder $\queryEncoder$ is used to translate the query $\querySeq$ into a fixed length representation of the same dimensionality as each \RUDAA{$\ctxTransducer_i(\documentSeq)$}. Both encoders use word embeddings computed by $e$ as their input. Then we compute a weight for every word in the document as the dot product of its contextual embedding and the query embedding. 
This weight might be viewed as an attention over the document $\documentSeq$. 

% In the end we normalize the weights to form a proper probability distribution over document's words by applying the $\text{softmax}$ function. 

To form a proper probability distribution over the words in the document, we normalize the weights using the \emph{softmax} function.
This way we model probability $s_i$ that the answer to query $\querySeq$ appears at position $i$ in the document $\documentSeq$.  
In a functional form this is:

\begin{equation}
%\queryEmbedded = \queryEncoder(\querySeq)
%\documentTransducedSeq \ctxTransducer(\documentSeq)
    s_i \propto \exp \left( \ctxTransducer_i(\documentSeq) \cdot \queryEncoder(\querySeq) \right)
    \label{eq:att}
\end{equation}

Finally we compute the probability that word $w$ is a correct answer as:

\begin{equation}
P(w|\querySeq,\documentSeq) \propto \sum_{i \in I(w,\documentSeq)} s_i
\label{eq:probSum}
\end{equation}

where $I(w,\documentSeq)$ is a set of positions where $w$ appears in the document $\documentSeq$.
%(it is often the case that one word appears at multiple locations in the document). 
We call this mechanism \textit{pointer sum attention} since we use attention as a pointer over discrete tokens in the context document and then we directly sum the word's attention across all the occurrences.
This differs from the usual use of attention in sequence-to-sequence models~\cite{Bahdanau2014} where attention is used to blend representations of words into a new embedding vector. 
Our use of attention was inspired by \glspl{ptrnet}~\cite{vinyals2015pointer}.

%Both $\ctxTransducer$ and $\queryEncoder$ share a single word embedding function $\embed$ that translates words into their word vectors. $\embed$ is trained together with the other parameters of the model. 
A high level structure of our model is shown in Figure~\ref{fig:model-structure}.

\subsection{Model instance details}
In our model the document encoder $\ctxTransducer$ is implemented as a bidirectional Gated Recurrent Unit (GRU) network~\cite{Cho2014,Chung2014} whose hidden states form the contextual word embeddings, that is $\ctxTransducer_i(\documentSeq) = \fwd{\ctxTransducer_i}(\documentSeq) \,\, ||\,\, \back{\ctxTransducer_i}(\documentSeq)$, where $||$ denotes vector concatenation and $\fwd{\ctxTransducer_i}$ and $\back{\ctxTransducer_i}$ denote forward and backward contextual embeddings from the respective recurrent networks.
The query encoder $\queryEncoder$ is implemented by another bidirectional GRU network. This time the last hidden state of the forward network is concatenated with the last hidden state of the backward network to form the query embedding, that is $\queryEncoder(\querySeq) = \fwd{\queryEncoder_{|\querySeq|}}(\querySeq)\,\, ||\,\, \back{\queryEncoder_1}(\querySeq)$.
The word embedding function $e$ is implemented in a usual way as a look-up table $\vocabMat$. $\vocabMat$ is a matrix whose rows can be indexed by words from the vocabulary, that is $e(w) = V_w , w \in V$. Therefore, each row of $\vocabMat$ contains embedding of one word from the vocabulary.  During training we jointly optimize parameters of $\ctxTransducer$, $\queryEncoder$ and $e$.
 %We use the usual setup where $e$ is realized by a simple look-up table.

%\[
%q_i = w_{i_1},w_{i_2} \dots w_{i_n}
%\]

\begin{figure}[t]
\small
\centering%
\begin{tabular}{| p{\dimexpr0.47\textwidth-2\tabcolsep-\arrayrulewidth\relax}|   }
\hline
\hspace{0.25\textwidth}\textbf{...}
\\
%according to a entity21 lawmaker (education policy is super gay, obviously); %\attvis{entity25}{0.86},
%who an op-ed writer for entity28, entity29, claims is being used by entity30 to 
%"attract young girls" to her show (uh-huh); the entity36 princess movie "entity40"
%according to radio hosts in entity38 (that dress!); and now, according to a 
%potential 2016 entity34 presidential contender, \attvis{entity32}{0.81}, there's prison. yep, %prison. 
%stay away from crime, kids. turns ya gay. \attvis{entity32}{0.705}, who ,let me reiterate , 
%is a potential presidential candidate from a major entity54 party 
what was supposed to be a fantasy sports car ride at \attvis{@entity3}{0.97} turned deadly when a \attvis{@entity4}{0.47} crashed into a guardrail . the crash took place sunday at the @entity8 , which bills itself as a chance to drive your dream car on a racetrack . the \attvis{@entity4}{.99} 's passenger , 36 - year - old @entity14 of @entity15 , @entity16 , died at the scene , @entity13 said . the driver of the \attvis{@entity4}{0.98} , 24 - year - old @entity18 of @entity19 , @entity16 , lost control of the vehicle , the \attvis{@entity13}{0.82} said .
\\
\hspace{0.25\textwidth}\textbf{...}
\\
\includegraphics[width=0.44\textwidth]{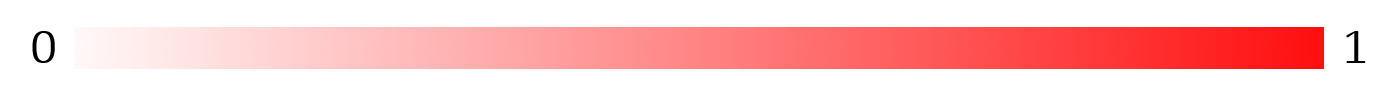}
\\
\hline
\hline
%\begin{figure}[h!]

% \caption{See that the ....}
% \label{fig:attention_1}
%\end{figure}

officials say the driver , 24 - year - old @entity18 , lost control of a \_\_\_\_\_
\\
\hline

\end{tabular}
\caption{Attention in an example with anonymized entities where our system selected the correct answer. Note that the attention is focused only on named entities.}
\label{tab:goodEx}
\end{figure}

\begin{figure}[t]
\small
\centering%
\begin{tabular}{| p{\dimexpr0.47\textwidth-2\tabcolsep-\arrayrulewidth\relax}|   }
\hline
\hspace{0.25\textwidth}\textbf{...}
\\
\attvis{@entity11}{0.444} film critic @entity29 writes in his review that "anyone 
nostalgic for childhood dreams of transformation will find something to enjoy in an uplifting movie 
that invests warm sentiment in universal themes of loss and resilience , experience and maturity . 
" more : the best and worst adaptations of "@entity" @entity43, \attvis{@entity44}{0.99} and @entity46 star 
in director @entity48's crime film about a hit man trying to save his estranged son from a revenge plot.
\attvis{@entity11}{0.873} chief film critic @entity52 writes in his review that the film 
\\
\hspace{0.25\textwidth}\textbf{...}
\\
\includegraphics[width=0.44\textwidth]{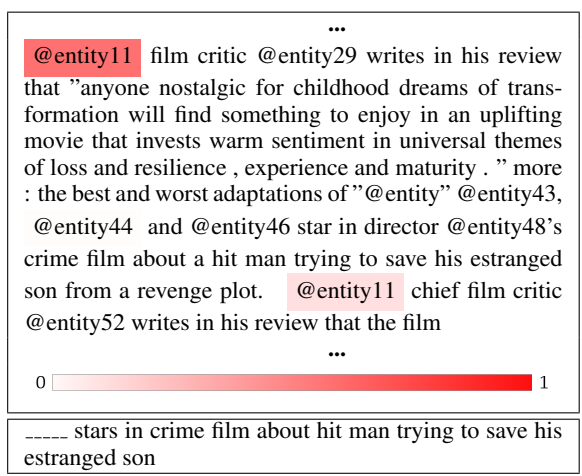}
\\
\hline
\hline
\_\_\_\_\_ stars in crime film about hit man trying to save his estranged son
\\
\hline
\end{tabular}
\caption{Attention over an example where our system failed to select the correct answer (entity43).
        The system was probably mislead by the co-occurring word 'film'.
        Namely, entity11 occurs $7$ times in the whole document and $6$ times it is together with the word 'film'.
        On the other hand, the correct answer occurs only $3$ times in total and only once together with 'film'.
        %Note also that attention is centered only on the named entity candidate.
        }
\end{figure}

\section{Related Work}
\RUDA{
Several recent deep neural network architectures~\cite{hermann2015teaching,hill2015goldilocks,chen2016thorough,Kobayashi2016} were applied to the task of text comprehension. The last two architectures were developed independently at the same time as our work. All of these architectures use an attention mechanism that allows them to highlight places in the document that might be relevant to answering the question. We will now briefly describe these architectures and compare them to our approach.

\subsection{Attentive and Impatient Readers}\label{subsec:attReaderComparison}
\emph{Attentive} and \emph{Impatient Readers} were proposed in~\cite{hermann2015teaching}. 
The simpler Attentive Reader is very similar to our architecture. 
It also uses bidirectional document and query encoders 
to compute an attention in a similar way we do. 
The more complex Impatient Reader computes attention over the document after reading every word of the query. 
However, empirical evaluation has shown that both models perform almost identically on the CNN and Daily Mail datasets.

%The key differences between Attentive reader and our model are: (i) Attentive reader computes attention in a slightly different way, e.g., it uses addition instead of dot product used in our model, (ii) attention is used to compute fixed length representation of $r$ of the document that is equal to weighted sum contextual embeddings of words in $\documentSeq$, that is $r = \sum_i s_i, \ctxTransducer_i(\documentSeq)$. 
The key difference between the Attentive Reader and our model is that the Attentive Reader uses attention to compute a fixed length representation $r$ of the document $\documentSeq$ that is equal to a weighted sum of contextual embeddings of words in $\documentSeq$, that is $r = \sum_i s_i \ctxTransducer_i(\documentSeq)$.
% \begin{equation}
% r = \sum_i s_i \ctxTransducer_i(\documentSeq)
% \end{equation}
A joint query and document embedding $m$ is then a non-linear function of $r$ and the query embedding $\queryEncoder(\querySeq)$. This joint embedding $m$ is in the end compared against all candidate answers $a' \in A$ using the dot product $\embed(a') \cdot m$, in the end the scores are normalized by $\text{softmax}$. That is: $P(a'|\querySeq,\documentSeq) \propto \exp \left(e(a') \cdot m \right)$. 
}

\MARTIN{
In contrast to the Attentive Reader, we select the answer from the context directly using the computed attention rather than using such attention for a weighted sum of the individual representations (see Eq.~\ref{eq:probSum}).
The motivation for such simplification is the following.

Consider a context ``\emph{A UFO was observed above our city in January and again in March.}'' and  question ``\emph{An observer has spotted a UFO in \_\_\_ .}''

Since both January and March are equally good candidates, the attention mechanism might put the same attention on both these candidates in the context. The blending mechanism described above would compute a vector between the representations of these two words and propose the closest word as the answer - this may well happen to be February (it is indeed the case for Word2Vec trained on Google News). 
By contrast, our model would correctly propose January or March. 
%It could very well be that the candidate with closest representation to the resulting blended embedding is Liverpool.%\footnote{It indeed is for word2vec vectors trained on Wikipedia.}.

% not present in the context document.
%\todo[inline]{Ruda: I am not sure whether this willbe a problem on CNN and DM because of anonymi}
}

\RUDAA{
\subsection{Chen et al. 2016}
A model presented in~\cite{chen2016thorough} is inspired by the Attentive Reader. One difference is that the attention weights are computed with a bilinear term instead of simple dot-product, that is $s_i \propto \exp \left( \ctxTransducer_i(\documentSeq)^\intercal\, W\, \queryEncoder(\querySeq) \right)$. The document embedding $r$ is computed using a weighted sum as in the Attentive Reader, $r = \sum_i s_i \ctxTransducer_i(\documentSeq)$. In the end $P(a'|\querySeq,\documentSeq) \propto \exp \left(e'(a') \cdot r \right)$, where $e'$ is a new embedding function.

Even though it is a simplification of the Attentive Reader this model performs significantly better than the original.

%Our model is even simpler since we base the prediction directly on $s_i$ attention values.

}

\MARTIN{
\subsection{Memory Networks}
\Glspl{MenNN}~\cite{weston2014memory} were applied to the task of text comprehension in~\cite{hill2015goldilocks}.

%\todo[inline]{intro k MemNN}

The best performing memory networks model setup - window memory - uses windows of fixed length (8) centered around the candidate words as memory cells.
Due to this limited context window, the model is unable to capture dependencies out of scope of this window.
Furthermore, the representation within such window is computed simply as the sum of embeddings of words in that window.
By contrast, in our model the representation of each individual word is computed using a recurrent network, which not only allows it to capture context from the entire document but also the embedding computation is much more flexible than a simple sum.

To improve on the initial accuracy, a heuristic approach called \textit{self supervision} is used in \cite{hill2015goldilocks} to help the network to select the right supporting ``memories'' using an attention mechanism showing similarities to the ours.
%\RUDAA{In the self-supervision phase the attention part of the network is trained to attend to the most promising occurrences of target words. This is similar to our training procedure.} 
Plain \glspl{MenNN} without this heuristic are not competitive on these machine reading tasks. Our model does not need any similar heuristics.

\RUDAA{
\subsection{Dynamic Entity Representation}
The Dynamic Entity Representation model~\cite{Kobayashi2016} has a complex architecture also based on the weighted attention mechanism and max-pooling over contextual embeddings of vectors for each named entity. %On CNN dataset the performance of a single model is comparable to the model reported in 

}

\subsection{Pointer Networks}
Our model architecture was inspired by \glspl{ptrnet}~\cite{vinyals2015pointer} in using an attention mechanism to select the answer in the context rather than to blend words from the context into an answer representation.
%  ??? The tasks this model is tailor-made to demonstrate the key differences.
%The tasks are sequence to sequence mapping problems, where the output dictionary is of variable size (planar convex hulls, Delaunay triangulations, planar Travelling Salesman Problem).
While a \gls{ptrnet} consists of an encoder as well as a decoder, which uses the attention to select the output at each step, our model outputs the answer in a single step.
Furthermore, the pointer networks assume that no input in the sequence appears more than once, which is not the case in our settings.
}

\subsection{Summary}
Our model combines the best features of the architectures mentioned above. 
We use recurrent networks to ``read'' the document and the query as 
\RUDAA{
done in~\cite{hermann2015teaching,chen2016thorough,Kobayashi2016}
}
%does Attentive and Impatient readers and we use attention in a way similar to \glspl{ptrnet}.
and we use attention in a way similar to \glspl{ptrnet}. 
We also use summation of attention weights in a way similar to \glspl{MenNN}~\cite{hill2015goldilocks}.

\RUDAA{
From a high level perspective we simplify all the discussed text comprehension models by removing all transformations past the attention step. Instead we use the attention directly to compute the answer probability.
}

%It should be noted that \glspl{MenNN}~\cite{hill2015goldilocks} also use summation of attention weights over individual ``memories'', however, they lack the power of recurrent neural networks.
%Compared to Attentive and Impatient readers and to \glspl{MenNN} our model is conceptually simpler. 

\section{Evaluation}

\begin{table*}[ht!]
\centering
  \begin{tabular}{@{}l@{}rr@{}l@{}rr@{}}
    \toprule
    & \multicolumn{2}{c}{CNN} &\phantom{aa}& \multicolumn{2}{c}{Daily Mail} \\
    \cmidrule{2-3} \cmidrule{5-6}
    & valid & test && valid & test \\
    \midrule
    %Maximum frequency              & 30.5 & 33.2 && 25.6 & 25.5 \\
    %Exclusive frequency            & 36.6 & 39.3 && 32.7 & 32.8 \\
    %Frame-semantic model~~         & 36.3 & 40.2 && 35.5 & 35.5 \\
    %Word distance model            & 50.5 & 50.9 && 56.4 & 55.5 \\
    %\midrule
    %Deep LSTM Reader \dg              & 55.0 & 57.0 && 63.3 & 62.2 \\
    %Uniform Reader                 & 39.0 & 39.4 && 34.6 & 34.4 \\
    Attentive Reader \dg             & 61.6 & 63.0 && 70.5 & 69.0 \\
    Impatient Reader \dg               & 61.8 & 63.8 && 69.0 & 68.0 \\
    \midrule
    MemNNs (single model) \ddg & 63.4 & 66.8 && NA & NA \\
    MemNNs (ensemble) \ddg    & 66.2 & 69.4 && NA & NA \\ 
    \midrule
    Dynamic Entity Repres. (max-pool) \dddg    & 71.2 & 70.7 && NA & NA \\ 
    Dynamic Entity Repres. (max-pool + byway)\dddg    & 70.8 & 72.0 && NA & NA \\ 
    Dynamic Entity Repres. + w2v \dddg    & 71.3 & 72.9 && NA & NA \\ 
    \midrule
    %Chen Classifier       & 64.2 & 65.9 && 67.5 & 66.6 \\ 
    Chen et al. (2016) (single model)       & 72.4 & 72.4 && 76.9 & 75.8 \\ 
    %Chen et al. (2016) (ensemble)      & 73.3 & 73.8 && NA & NA \\ 
    \midrule

    % Using our accuracy computation (after loading a saved model)
    \gls{psr} (single model)           & 68.6 & 69.5 && 75.0 & 73.9 \\ % New DM: 75.0 & 73.9
    \gls{psr} (avg for top 20\%)       & 68.4 & 69.9 && 74.5 & 73.5 \\
    \bf{\gls{psr} (avg ensemble)}   & 73.9 & \bf 75.4 && 78.1 & 77.1 \\ %New DM: 77.2 & 76.2
    \bf{\gls{psr} (greedy ensemble)}   & 74.5 &  74.8 && 78.7 & \bf 77.7 \\
    \bottomrule
  \end{tabular}
\caption{Results of our \gls{psr} on the CNN and Daily Mail datasets. Results for models marked with \dg{} are taken from~\protect\cite{hermann2015teaching}, results of models marked with \ddg{} are taken
from~\protect\cite{hill2015goldilocks} and results marked with \dddg{} are taken
from~\protect\cite{Kobayashi2016}. Performance of \ddg{} and \dddg{} models was evaluated only on CNN dataset.}
 \label{tab:results-cnn+dm}

\end{table*}

\begin{table*}[th!]
\centering
  \begin{tabular}{@{}l@{}rr@{}l@{}rr@{}}
    \toprule
    & \multicolumn{2}{c}{Named entity} &\phantom{aa}& \multicolumn{2}{c}{Common noun} \\
    \cmidrule{2-3} \cmidrule{5-6}
    & valid & test && valid & test \\
    \midrule
        Humans (query) $^{(*)}$ & NA & 52.0 &&  NA & 64.4 \\ %& 0.716 & 0.676 \\
        Humans (context+query) $^{(*)}$ &NA &{\it \textbf{81.6}} && NA & {\it \textbf{ 81.6}} \\ %& {\it \textbf{0.828}} & 0.708 \\
        \midrule
        LSTMs (context+query) \ddg & 51.2 & 41.8 && 62.6 & 56.0 \\ %& \bf 0.818 & 0.791 \\
        \midrule
        MemNNs  (window memory + self-sup.) \ddg & 70.4 & 66.6 &&  64.2 &  63.0 \\ %0.690 & 0.703\\
        \midrule
        \gls{psr} (single model)     & 73.8 & 68.6 && 68.8 & 63.4 \\ %    
        \gls{psr} (avg for top 20\%)     & 73.3 & 68.4 && 67.7 & 63.2 \\ %    NE_-b_32_-qice_True_-ehd_256_-sed_300_-lr_0.001.out
        \bf{\gls{psr} (avg ensemble)}     & 74.5 &  70.6 && 71.1 & \bf 68.9 \\
        \bf{\gls{psr} (greedy ensemble)}     &  76.2 & \bf 71.0 &&  72.4 &  67.5 \\
     \bottomrule
  \end{tabular}
\caption{Results of our \gls{psr} on the CBT datasets. Results marked with \ddg{} are taken
from~\protect\cite{hill2015goldilocks}. $^{(*)}$Human results were collected on 10\% of the test set.}
\label{tab:results-cbt}
\end{table*}

In this section we evaluate our model on the CNN, Daily Mail and CBT datasets. We show that despite the model's simplicity its ensembles achieve state-of-the-art performance on each of these datasets.

\subsection{Training Details}

\RUDA{

To train the model we used stochastic gradient descent with the ADAM update rule~\cite{Kingma2015} and learning rate of $0.001$ or $0.0005$. During training we minimized the following negative log-likelihood with respect to $\theta$:

\begin{equation}
- log P_\theta(a|\querySeq,\documentSeq)
\label{eq:objective}
\end{equation}
where $a$ is the correct answer for query $\querySeq$ and document $\documentSeq$,  and $\theta$ represents parameters of the encoder functions $\ctxTransducer$ and $\queryEncoder$ and of the word embedding function $e$. The optimized probability distribution $P(a|\querySeq,\documentSeq)$ is defined in Eq.~\ref{eq:probSum}.

The initial weights in the word embedding matrix were drawn randomly uniformly from the interval $[-0.1, 0.1]$. Weights in the GRU networks were initialized by random orthogonal matrices~\cite{Saxe2014} and biases were initialized to zero. We also used a gradient clipping~\cite{Pascanu2012} threshold of 10 and batches of size 32.

During training we randomly shuffled all examples in each epoch. To speedup training, we always pre-fetched $10$ batches worth of examples and sorted them according to document length. Hence each batch contained documents of roughly the same length.

For each batch of the CNN and Daily Mail datasets we randomly reshuffled the assignment of named entities to the corresponding word embedding vectors to match the procedure proposed in~\cite{hermann2015teaching}. This guaranteed that word embeddings of named entities were used only as semantically meaningless labels not encoding any intrinsic features of the represented entities. This forced the model to truly deduce the answer from the single context document associated with the question.
}
\RUDAA{
We also do not use pre-trained word embeddings to make our training procedure comparable to~\cite{hermann2015teaching}.
}

We did not perform any text pre-processing since the original datasets were already tokenized.

\RUDAA{
We do not use any regularization since in our experience it leads to longer training times of single models, however, performance of a model ensemble is usually the same. This way we can train the whole ensemble faster when using multiple GPUs for parallel training.
}

\RUDAA{
For Additional details about the training procedure see Appendix~\ref{app:train}.
}

\ONDRA{
%During training we saved the model and computed validation accuracy at the end of each epoch and subsequently used the instance of the model with best validation performance. 

\subsection{Evaluation Method}

We evaluated the proposed model both as a single model and using ensemble averaging. 
\RUDAA{Although the model computes attention for every word in the document we restrict the model to select an answer from a list of candidate answers associated with each question-document pair.}

For single models we are reporting results for the best model as well as the average of accuracies for the best 20\% of models with best performance on validation data since single models display considerable variation of results due to random weight initialization\footnote{The standard deviation for models with the same hyperparameters was between 0.6-2.5\% in absolute test accuracy.} even for identical hyperparameter values. Single model performance may consequently prove difficult to reproduce.
}

What concerns ensembles, we used simple averaging of the answer probabilities predicted by ensemble members. %\footnote{Attempts to use the \emph{Constrained Optimization By Linear Approximation} (COBYLA) method \cite{Powell1994} to optimize the weights lead to overfitting on the validation data with respect to which the optimization was done.}.  
For ensembling we used 14, 16, 84 and 53 models for CNN, Daily Mail and CBT CN and NE respectively. The ensemble models were chosen either as the top 70\% of all trained models, we call this \textit{avg ensemble}.
Alternatively we use the following algorithm:
We started with the best performing model according to validation performance. Then in each step we tried adding the best performing model that had not been previously tried. We kept it in the ensemble if it did improve its validation performance and discarded it otherwise. This way we gradually tried each model once. We call the resulting model a \textit{greedy ensemble}.

 \begin{figure*} [hpt]
        \label{fig:contextLength}
        \centering
        \begin{subfigure}[b]{0.475\textwidth}
            \centering
            \includegraphics[width=\textwidth]{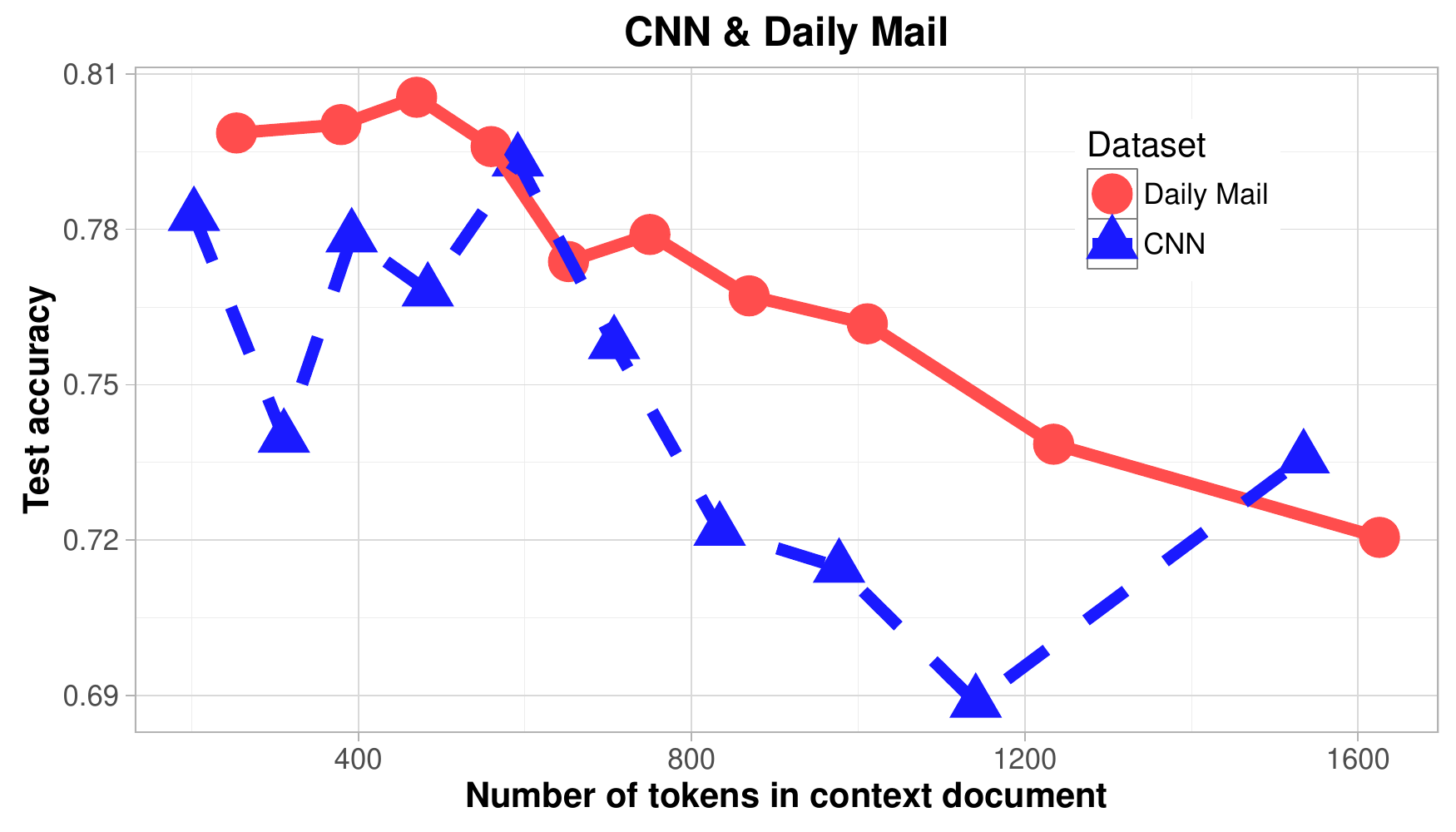}
            \caption[]%
            {{}}    
            \label{fig:contextLength_cnn_dm}
        \end{subfigure}
        \hfill
        \begin{subfigure}[b]{0.475\textwidth}  
            \centering 
            \includegraphics[width=\textwidth]{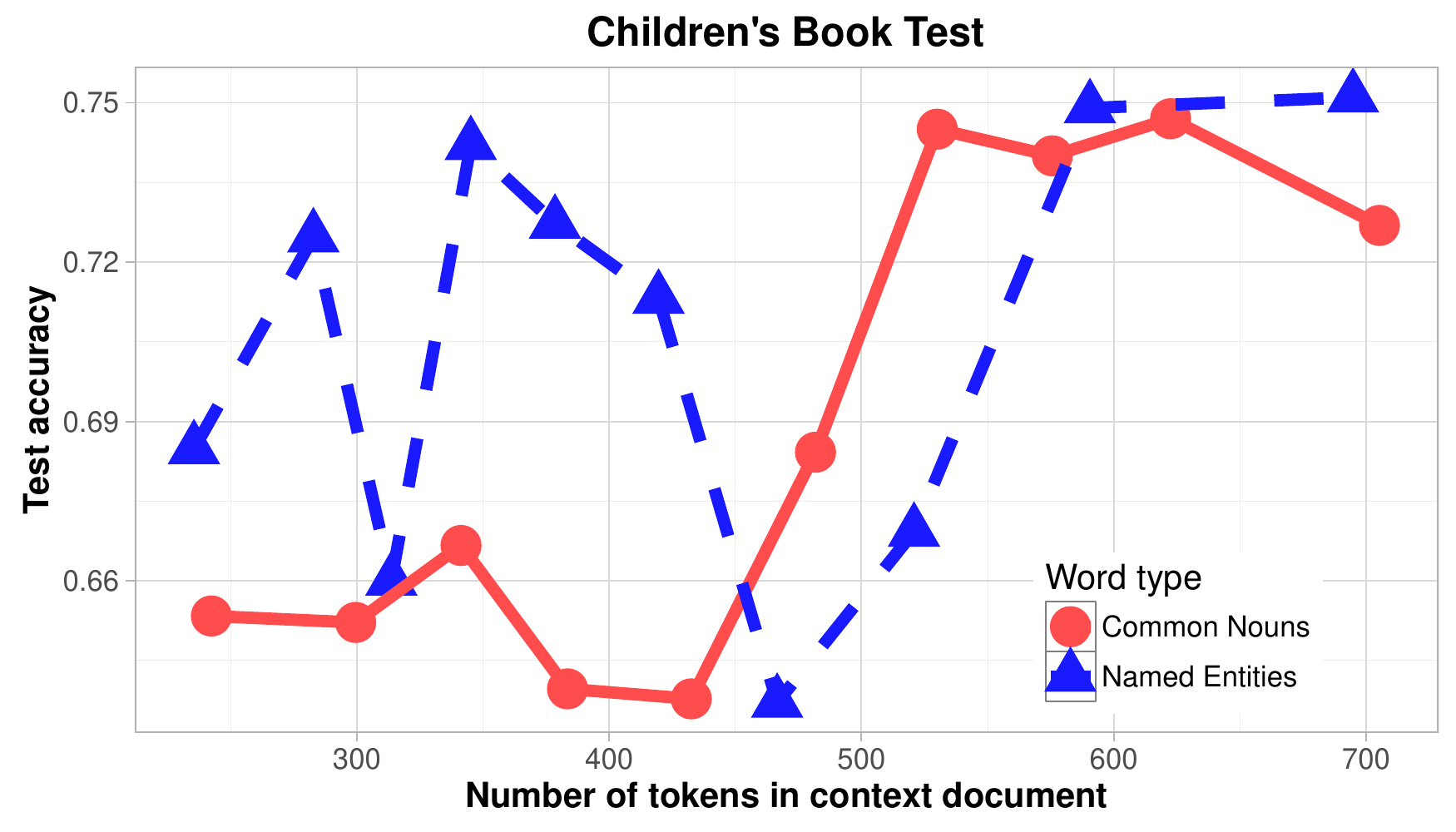}
            \caption[]%
            {{}}    
            \label{fig:contextLength_cnn_dm_hist}
        \end{subfigure}
        \vskip\baselineskip
        \begin{subfigure}[b]{0.475\textwidth}   
            \centering 
            \includegraphics[width=\textwidth]{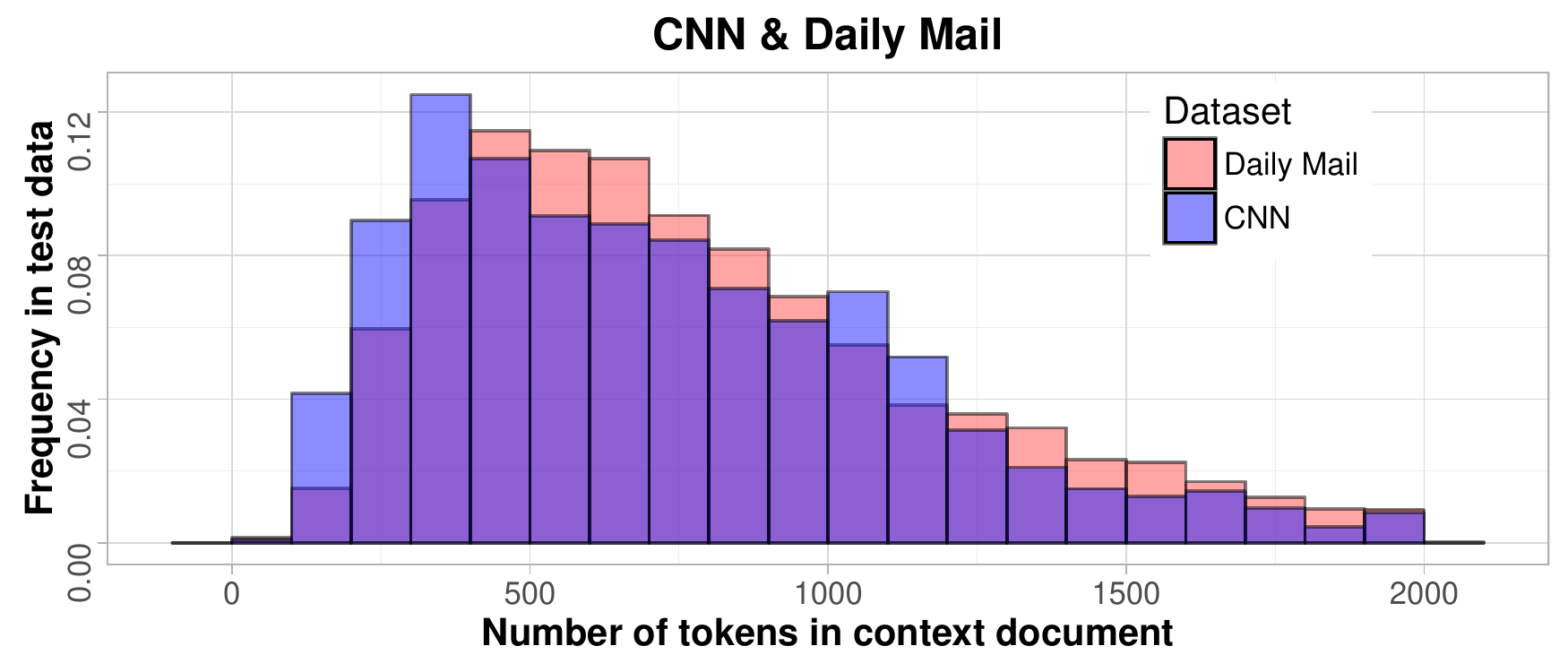}
            \caption[]%
            {{}}    
            \label{fig:contextLength_cbt}
        \end{subfigure}
        \quad
        \begin{subfigure}[b]{0.475\textwidth}   
            \centering 
            \includegraphics[width=\textwidth]{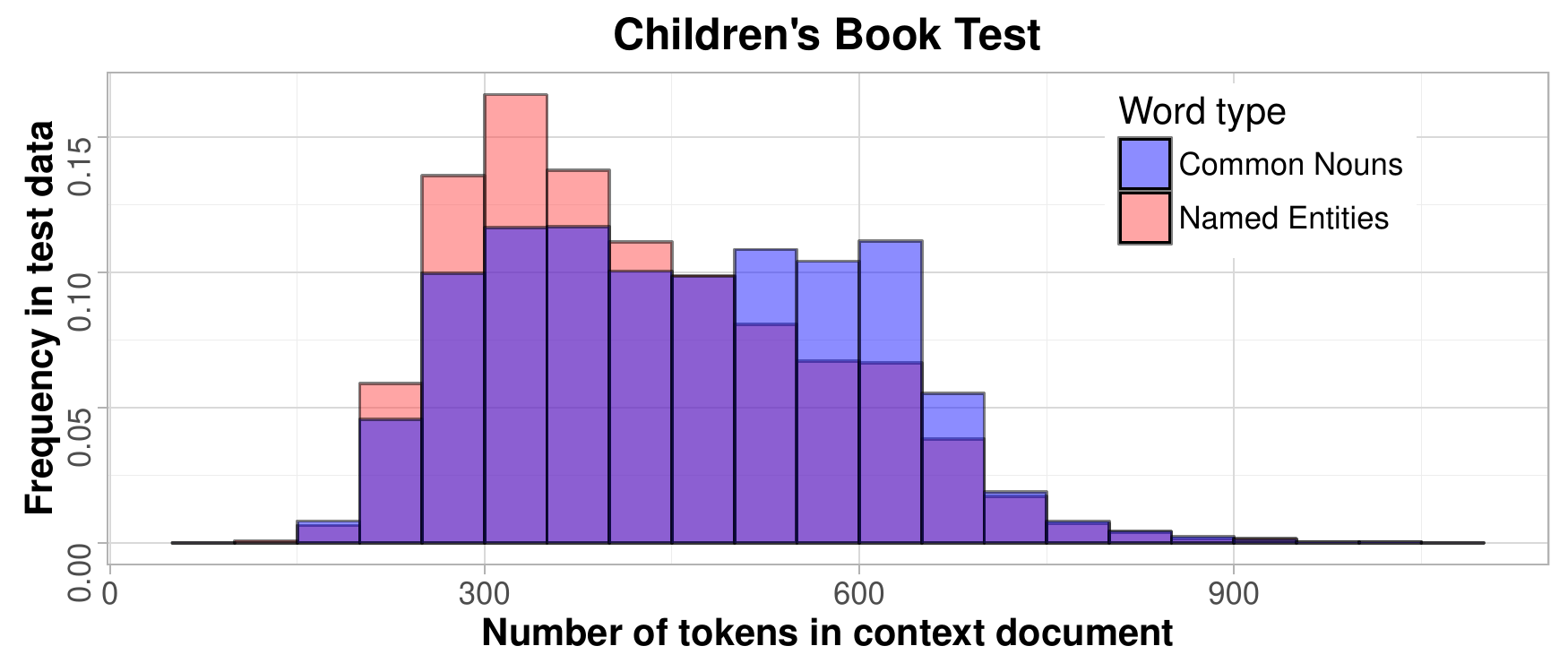}
            \caption[]%
            {{}}    
            \label{fig:contextLength_cbt_hist}
        \end{subfigure}
        \caption[ The average and standard deviation of critical parameters ]{
        %Subfigures~\ref{fig:cnn_acc_by_ctx} and~\ref{fig:dm_acc_by_ctx} show how accuracy depends on context length.}
        Sub-figures (a) and (b) plot the test accuracy against the length of the context document. The examples were split into ten buckets of equal size by their context length. Averages for each bucket are plotted on each axis. Sub-figures (c) and (d) show distributions of context lengths in the four datasets.}
        \label{fig:cnn+dm_lengthAcc}
\end{figure*}

\begin{figure} [hpb]
        \centering
        \begin{subfigure}[b]{0.475\textwidth}
            \centering
            \includegraphics[width=\textwidth]{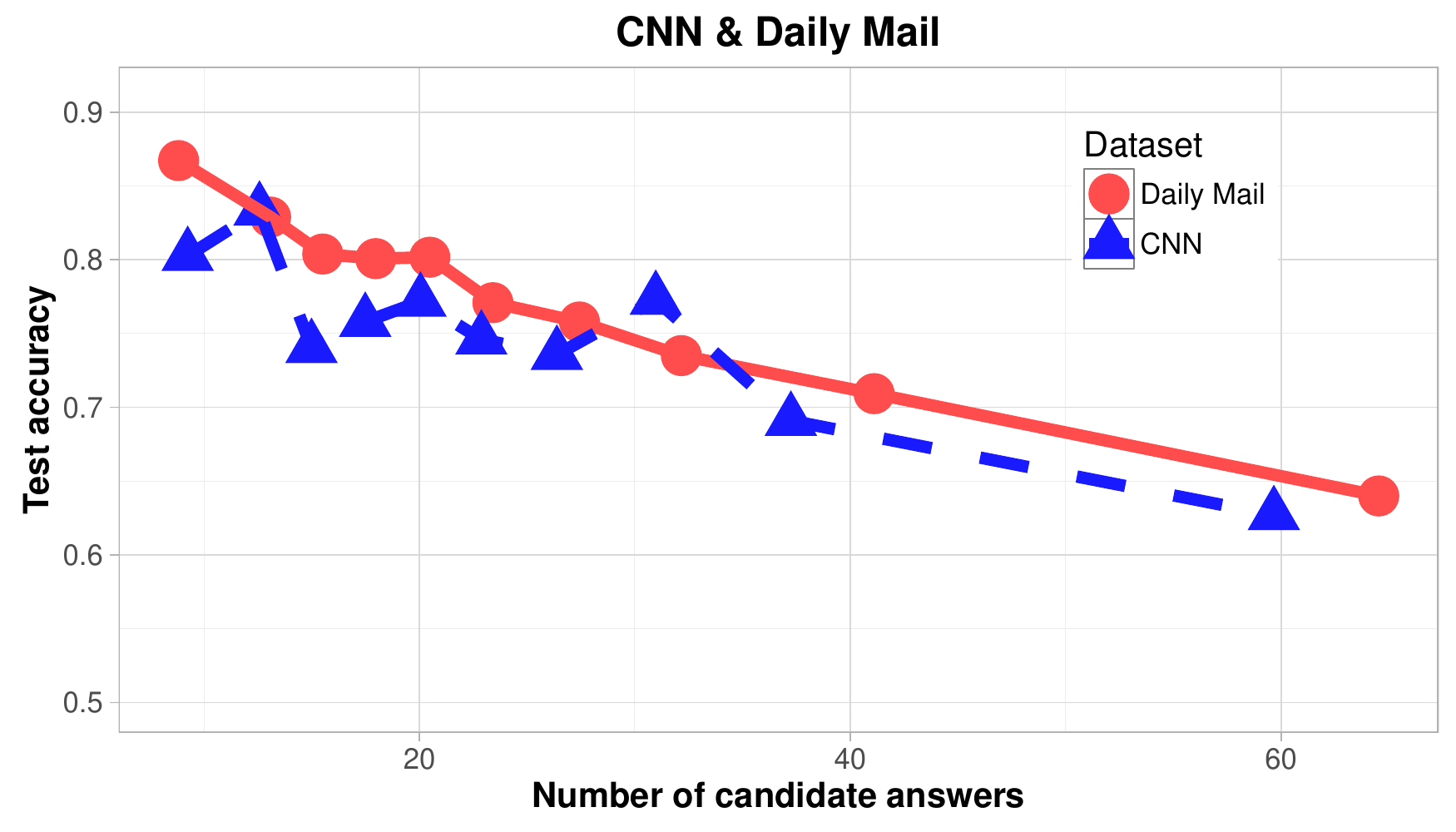}
            \caption[]%
            {{}}    
            \label{fig:cnn+dm_candidateEntities_graph}
        \end{subfigure}

        \vskip\baselineskip
        \begin{subfigure}[b]{0.475\textwidth}   
            \centering 
            \includegraphics[width=\textwidth]{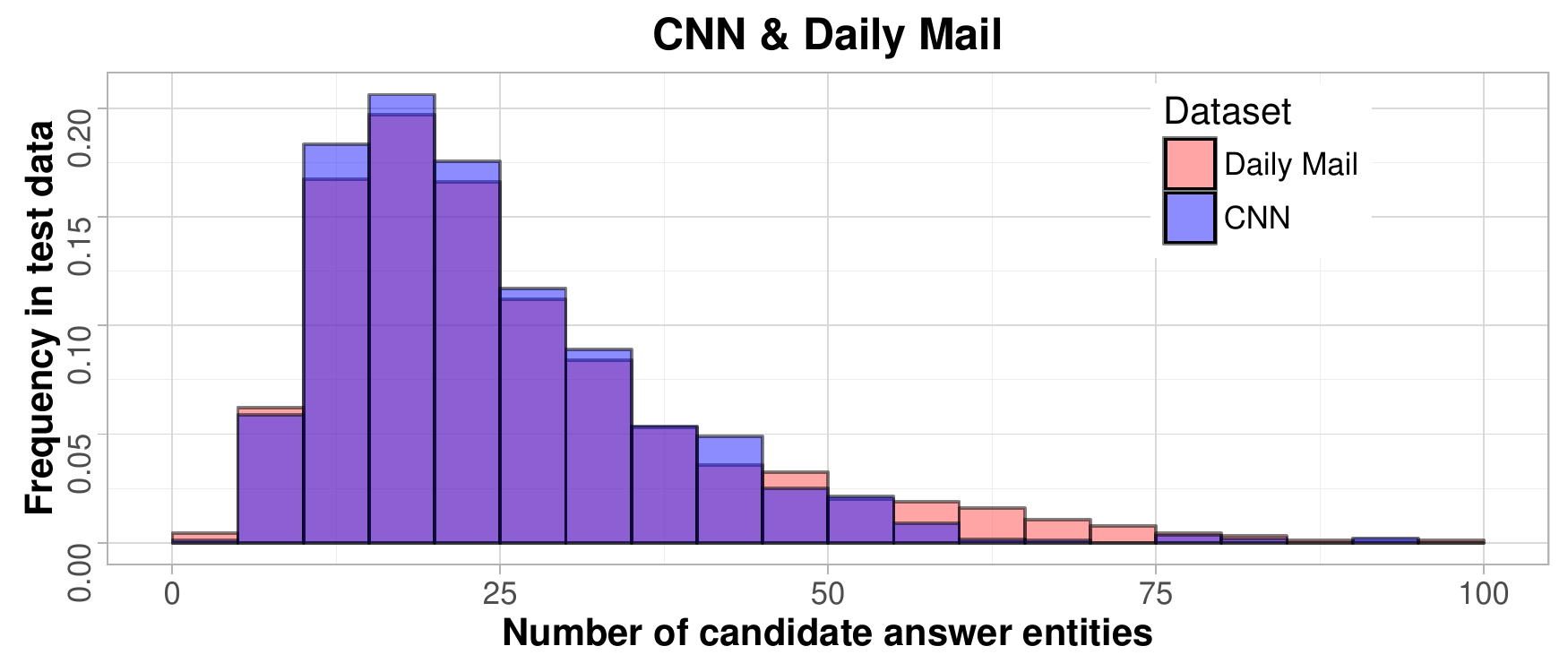}
            \caption[]%
            {{}}
            \label{fig:cnn+dm_candidateEntities_hist}
        \end{subfigure}
        \caption{
       Subfigure (a) illustrates how the model accuracy decreases with an increasing number of candidate named entities. Subfigure (b) shows the overall distribution of the number of candidate answers in the news datasets.}
        \label{fig:cnn+dm_candidateEntities}
\end{figure}

\begin{figure} [hpb]
        \centering
        \begin{subfigure}[b]{0.475\textwidth}
            \centering
            \includegraphics[width=\textwidth]{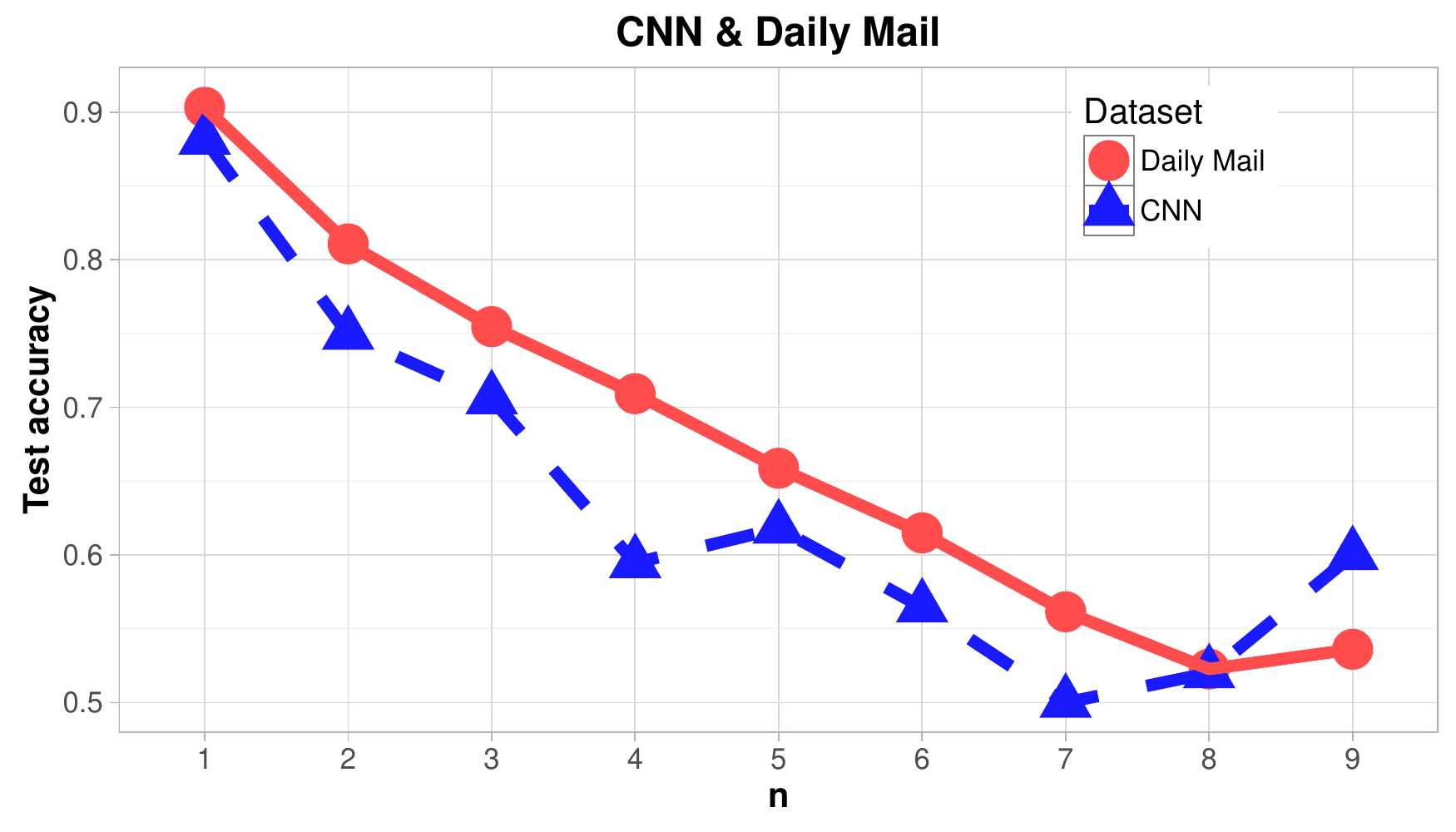}
            \caption[]%
            {{}}    
            \label{fig:cnn+dm_nMostFreq_graph}
        \end{subfigure}

        \vskip\baselineskip
        \begin{subfigure}[b]{0.475\textwidth}   
            \centering 
            \includegraphics[width=\textwidth]{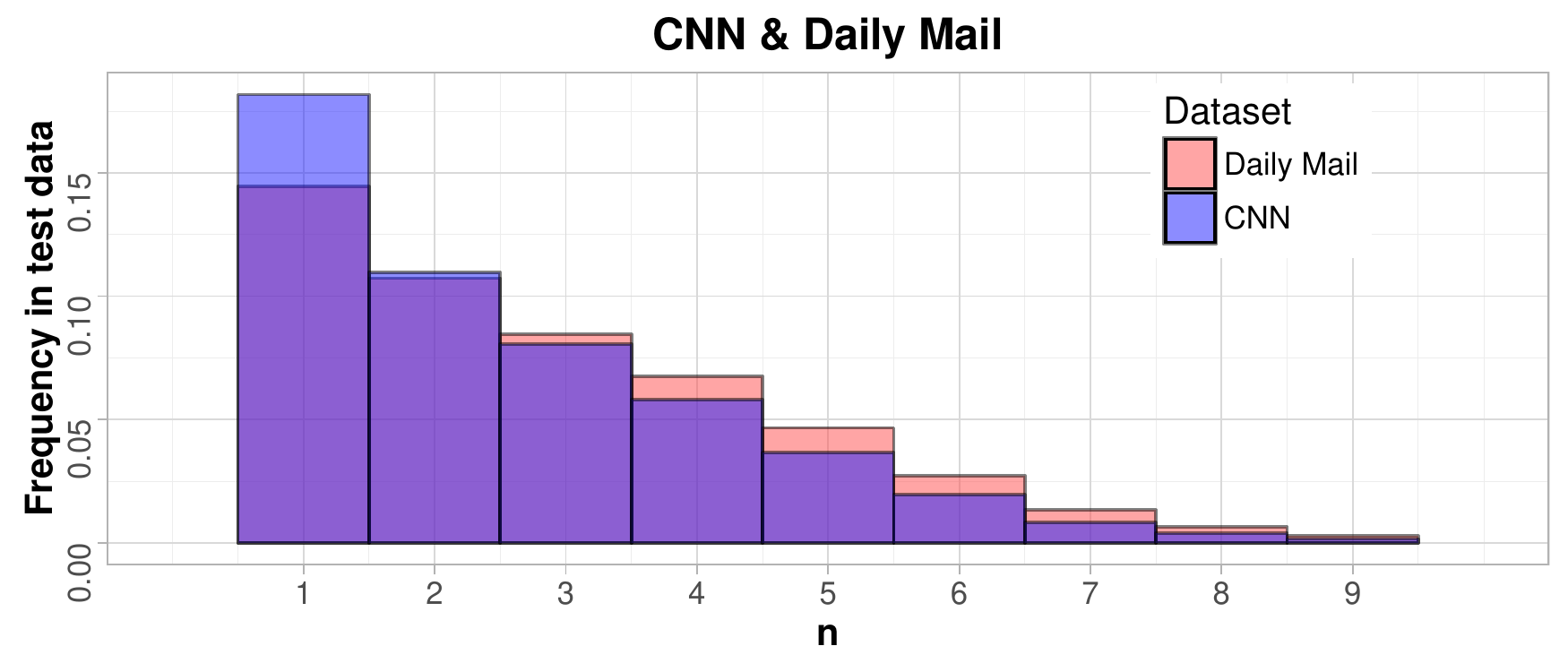}
            \caption[]%
            {{}}    
            \label{fig:cnn+dm_nMostFreq_hist}
        \end{subfigure}
        \caption{
       Subfigure (a) shows the model accuracy when the correct answer is the $n^{\text{th}}$ most frequent named entity for $n\in [1,10]$. Subfigure (b) shows the number of test examples for which the correct answer was the $n$--th most frequent entity. The plots for CBT look almost identical (see Appendix~\ref{app:A}).}%very similar and are provided in the appendix.}
        \label{fig:cnn+dm_nMostFreq}
\end{figure}

\subsection{Results}

\RUDA{
Performance of our models on the CNN and Daily Mail datasets is summarized in Table~\ref{tab:results-cnn+dm}, Table~\ref{tab:results-cbt} shows results on the CBT dataset.
%CBT dataset is summarized in Table~\ref{tab:results-cbt}, Table~\ref{tab:results-cnn+dm} shows results on the CNN and Daily Mail datasets. 
The tables also list performance of other published models that were evaluated on these datasets. 
}
Ensembles of our models set new state-of-the-art results on all evaluated datasets. 

%\todo[inline]{DEL: However, even our best single models outperform the best previously reported results.}

Table~\ref{tab:results-nBest} then measures accuracy as the proportion of test cases where the ground truth was among the top $k$ answers proposed by the greedy ensemble model for $k=1,2,5$.

\textbf{CNN and Daily Mail.} 
\RUDAA{The CNN dataset is the most widely used dataset for evaluation of text comprehension systems published so far. Performance of our single model is a little bit worse than performance of simultaneously published models ~\cite{chen2016thorough,Kobayashi2016}. Compared to our work these models were trained with Dropout regularization~\cite{Srivastava2014} which might improve single model performance. However, ensemble of our models outperforms these models even though they use pre-trained word embeddings.

%The best single model selected by validation accuracy reported by ~\cite{Kobayashi2016} that does not use  pre-trained word embeddings achieves test accuracy 70.7\% (1.2\% better than our single best model). The best model with pre-trained embeddings is 3.4\% better than our single model. The models reported in~\cite{chen2016thorough} use pretrained embeddings.

}

On the CNN dataset our single model with best validation accuracy achieves a test accuracy of 69.5\%. The average performance of the top 20\% models according to validation accuracy is 69.9\% which is even 0.5\% better than the single best-validation model. This shows that there were many models that performed better on test set than the best-validation model. Fusing multiple models then gives a significant further increase in accuracy on both CNN and Daily Mail datasets..% Our simple-average ensemble outperforms the accuracy of the best previously reported ensemble~\cite{hill2015goldilocks} by 6\%.

%On the Daily Mail dataset our averaging ensemble achieves performance of 77.1\%. 
%single model with accuracy of 73.9\% outperforms the best previous result achieved by the Attentive Reader~\cite{hermann2015teaching} by 4.9\% absolute and our averaging ensemble is by 8.7\% absolute better. 

\textbf{CBT.} In named entity prediction our best single model with accuracy of 68.6\% performs 2\% absolute better than the \gls{MenNN} with self supervision, the averaging ensemble performs 4\% absolute better than the best previous result. In common noun prediction our single models is 0.4\% absolute better than \gls{MenNN} however the ensemble improves the performance to 69\% which is 6\% absolute better than \gls{MenNN}.
%When we analyze performance of the other models that performed well 

% \begin{figure*}[ht!] \label{ fig7} 
%   \begin{minipage}[b]{0.5\linewidth}
%      \includegraphics[width=1\linewidth]{cnn_contextLength_accuracy.pdf} 
%      \caption{The dependence of the accuracy on the length of context in CNN.} 
%   \end{minipage} 
%   \begin{minipage}[b]{0.5\linewidth}
%      \includegraphics[width=1\linewidth]{DM_contextLength_accuracy.pdf} 
%      \caption{Rupture} 
%   \end{minipage} 
%   \begin{minipage}[b]{0.5\linewidth}
%     \includegraphics[width=1\linewidth]{CNN_contextTokens_hist.pdf} 
%     \caption{DFT, Initial condition} 
%   \end{minipage}
%   \hfill
%   \begin{minipage}[b]{0.5\linewidth}
%     \includegraphics[width=1\linewidth]{DM_contextTokens_hist.pdf}} 
%     \caption{DFT, rupture} 
%   \end{minipage} 
%   \caption{ABC}
% \end{figure*}

\begin{table}[h]
\centering

  \begin{tabular}{l|ccc}%{@{}l@{}rr@{}l@{}rr@{}r@{}}
    \toprule
    Dataset & $k=1$ & $k=2$ & $k=5$\\
    %\cmidrule{2-3} \cmidrule{5-6}
    %& valid & test && valid & test \\
    
    \midrule
    
        CNN & 74.8 & 85.5 & 94.8\\
        Daily Mail & 77.7 & 87.6 & 94.8\\
        CBT NE & 71.0 & 86.9 & 96.8 \\
        CBT CN& 67.5 & 82.5 & 95.4 \\ 
    
     \bottomrule
  \end{tabular}
\caption{Proportion of test examples for which the top $k$ answers proposed by the greedy ensemble included the correct answer.}
\label{tab:results-nBest}
\end{table}

\ONDRA{
\section{Analysis}
\label{sec:analysis}

%\subsection{Influence of task properties}

To further analyze the properties of our model, we examined the dependence of accuracy on the length of the context document (Figure~\ref{fig:cnn+dm_lengthAcc}), the number of candidate answers (Figure~\ref{fig:cnn+dm_candidateEntities}) and the frequency of the correct answer in the context (Figure~\ref{fig:cnn+dm_nMostFreq}).

On the CNN and Daily Mail  datasets, the accuracy decreases with increasing document length (Figure \ref{fig:contextLength_cnn_dm}). 
We hypothesize this may be due to multiple factors. 
Firstly long documents may make the task more complex. 
Secondly such cases are quite rare in the training data (Figure~\ref{fig:contextLength_cnn_dm_hist}) which motivates the model to specialize on shorter contexts. 
Finally the context length is correlated with the number of named entities, i.e. the number of possible answers which is itself negatively correlated with accuracy (see Figure~\ref{fig:cnn+dm_candidateEntities}).

%On the CBT dataset, there seems to be no dependence on context length (\ref{fig:contextLength_cbt}).
On the CBT dataset this negative trend seems to disappear (Fig. \ref{fig:contextLength_cbt}).
This supports the later two explanations since the distribution of document lengths is somewhat more uniform (Figure~\ref{fig:contextLength_cbt_hist})
and the number of candidate answers is constant ($10$) for all examples in this dataset. 
%However further investigation is needed to specify the contributions of the individual factors.

The effect of increasing number of candidate answers on the model's accuracy can be seen in Figure~\ref{fig:cnn+dm_candidateEntities_graph}.
We can clearly see that as the number of candidate answers increases, the accuracy drops. 
On the other hand, the amount of examples with large number of candidate answers is quite small (Figure~\ref{fig:cnn+dm_candidateEntities_hist}).

Finally, since the summation of attention in our model inherently favours frequently occurring tokens, we also visualize how the accuracy depends on the frequency of the correct answer in the document.
Figure~\ref{fig:cnn+dm_nMostFreq_graph} shows that the accuracy significantly drops as the correct answer gets less and less frequent in the document compared to other candidate answers.
On the other hand, the correct answer is likely to occur frequently (Fig. \ref{fig:cnn+dm_nMostFreq_graph}).

% We also looked on the dependence of accuracy on the how frequent the correct answer was in the context document - since the summation of attention in our model inherently favours frequently occurring tokens we expected that the model would perform worse on answers rare in the context. This was indeed confirmed by our analysis as can be seen in Figure~\ref{fig:cnn+dm_nMostFreq}.
}

%\subsection{Comparison to Weighted Average Blending}
%We hypothesized in Section~\ref{subsec:attReaderComparison} that the fact that the Attentive Reader uses attention to create a blended representation potentially harms its performance.
%In order to verify this intuition, we implemented blending into our model, bringing it closer to the Attentive Reader \cite{hermann2015teaching}.

%In this modified model we compute attention weights $s_i$ in the same way as in our original model (see Eq.~\ref{eq:att}). However, we replace Eq.~\ref{eq:probSum} with the following equations:
%\begin{equation}
%    r = \sum_i s_i e(w_i)
%\end{equation}

%\begin{equation}
%    P(a'|\querySeq,\documentSeq) \propto \exp \left( r \cdot e(a')\right) 
%\end{equation}
%where $r$ is the blended response embedding and $a' \in A$ is a possible candidate response. This change in the architecture indeed lead to a significant decrease in accuracy across all four datasets. 

%Namely, on each CBT dataset the difference was almost 15\% while on CNN and Daily Mail the decrease was over 6\% and 2\% respectively. Besides the training time for the models with blending was several times longer than our attention sum architecture both measured by the time per epoch and by the number of epochs required for the model to converge.

\RUDA{

\section{Conclusion}
In this article we presented a new neural network architecture for natural language text comprehension. 
While our model is simpler than previously published models, it gives a new state-of-the-art accuracy on all evaluated datasets.

An analysis by~\cite{chen2016thorough} suggests that on CNN and Daily Mail datasets a significant proportion of questions is ambiguous or too difficult to answer even for humans (partly due to entity anonymization) so the ensemble of our models may be very near to the maximal accuracy achievable on these datasets.

%This model is simpler than previously published models - instead of calculating a blended representation of an ideal answer, it directly sums the attention over each unique word occurring in the context document. 
%Our model gives a new state-of-the-art accuracy on the CNN, Daily Mail and Children's Book Test Datasets and we provided some evidence that it may be due to the above mentioned simplification. 

%In this work we presented a new neural network architecture for text comprehension. The architecture is simpler than models known from previously published literature, such as the Attentive Reader~\cite{hermann2015teaching} or \glspl{MenNN}~\cite{hill2015goldilocks}, however, it achieves a state-of-the-art performance on all large scale text comprehension datasets~\cite{hermann2015teaching,hill2015goldilocks} that are suitable for deep learning models.

% The model performance may be further improved by pre-training the source embedding on additional text corpora, by using larger training datasets\footnote{For instance simply combining the CNN and Daily Mail datasets leads to a further 1-2% improvement in accuracy over models trained only on the Daily Mail data.} or by combining our model focusing on the identification of a fictional

%Future experiments might investigate whether our model can take advantage of multi-step inference implemented in neural networks~\cite{weston2014memory,Sukhbaatar2015,Kumar2015,Yu2015a} that allows the model to reason about the text in more detail.  

}

\section*{Acknowledgments}

We would like to thank Tim Klinger for providing us with masked softmax code that we used in our implementation.

\bibliography{citations}
\bibliographystyle{acl2016}

\newpage
\begin{appendices}

\section{Training Details}
\label{app:train}
During training we evaluated the model performance after each epoch and stopped the training when the error on the validation set started increasing. %After a single epoch when the validation accuracy}

The models usually converged after two epochs of training. Time needed to complete a single epoch of training on each dataset on an Nvidia K40 GPU is shown in Table~\ref{tab:timing}.

\begin{table}[ht]
\centering
  \begin{tabular}{lr@{~}l}
    \toprule
    Dataset & \multicolumn{2}{l}{Time per epoch} \\
    %\cmidrule{2-3} \cmidrule{5-6}
    %& valid & test && valid & test \\
    
    \midrule
    
        CNN & 10h &22min \\
        Daily Mail & 25h &42min \\
        CBT Named Entity & ~1h & ~~5min \\
        CBT Common Noun & ~0h &56min \\ 
    
     \bottomrule
  \end{tabular}
  \caption{Average duration of one epoch of training on the four datasets.}
  \label{tab:timing}

\end{table}

\ONDRA{
The hyperparameters, namely the recurrent hidden layer dimension and the source embedding dimension, were chosen by grid search. We started with a range of 128 to 384 for both parameters and subsequently kept increasing the upper bound by 128 until we started observing a consistent decrease in validation accuracy. The region of the parameter space that we explored together with the parameters of the model with best validation accuracy are summarized in Table~\ref{tab:params}. 
}

\begin{table}[ht]
\centering
  \resizebox{0.45\textwidth}{!}{
  \begin{tabular}{l|lll|lll}%{@{}l@{}rr@{}l@{}rr@{}r@{}}
    \toprule
     & \multicolumn{3}{c}{Rec. Hid. Layer} & \multicolumn{3}{c}{Embedding} \\
    Dataset & min & max & best & min & max & best\\
    %\cmidrule{2-3} \cmidrule{5-6}
    %& valid & test && valid & test \\
    
    \midrule
    
        CNN & 128 & 512 & 384 & 128 & 512 & 128 \\
        Daily Mail & 128 & 1024 & 512 & 128 & 512 & 384 \\
        CBT NE & 128 & 512 &384 & 128& 512 & 384 \\
        CBT CN& 128 & 1536 & 256 & 128 & 512 & 384 \\ 
     \bottomrule
  \end{tabular}
  }
  \caption{Dimension of the recurrent hidden layer and of the source embedding for the best model and the range of values that we tested. \RUDAA{We report number of hidden units of the unidirectional GRU; the bidirectional GRU has twice as many hidden units.}}
  \label{tab:params}

\end{table}

\RUDA{
Our model was implemented using Theano~\cite{Bastien-Theano-2012} and Blocks~\cite{VanMerrienboer2015}.
}

\section{Dependence of accuracy on the frequency of the correct answer}
\label{app:A}
In Section~\ref{sec:analysis} we analysed how the test accuracy depends on how frequent the correct answer is compared to other answer candidates for the news datasets. The plots for the Children's Book Test looks very similar, however we are adding it here for completeness.

\begin{figure} [hpb]
        \centering
        \begin{subfigure}[b]{0.475\textwidth}
            \centering
            \includegraphics[width=\textwidth]{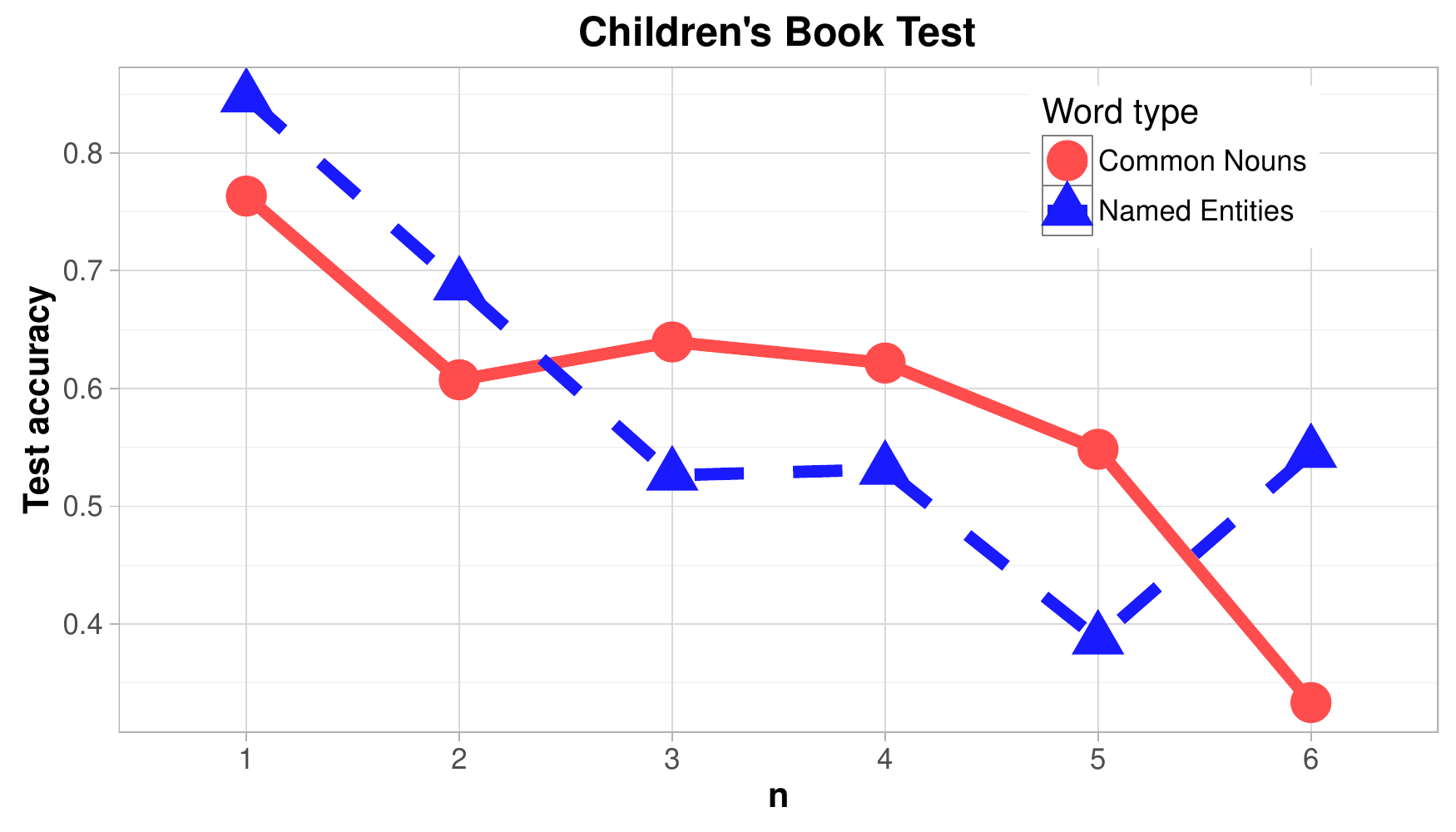}
            \caption[]%
            {{}}    
            \label{fig:cbt_nMostFreq_graph}
        \end{subfigure}

        \vskip\baselineskip
        \begin{subfigure}[b]{0.475\textwidth}   
            \centering 
            \includegraphics[width=\textwidth]{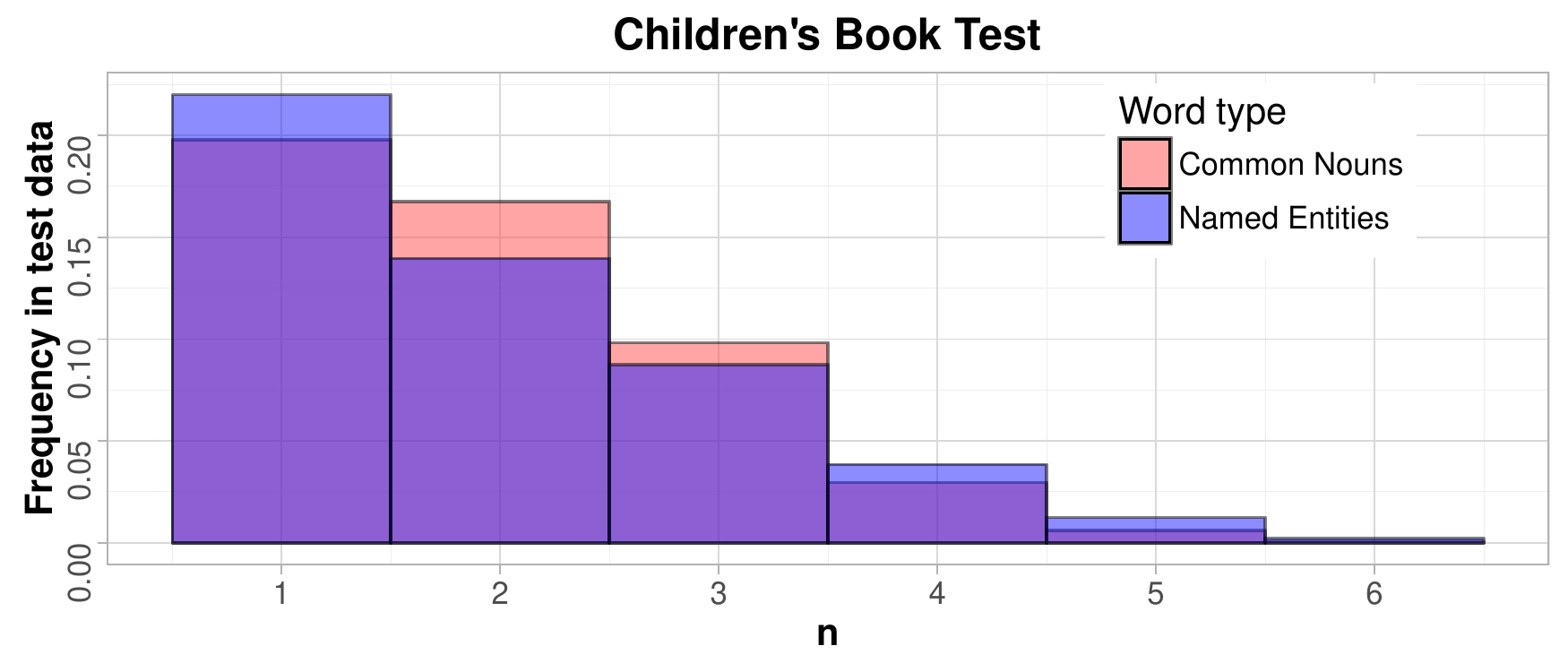}
            \caption[]%
            {{}}    
            \label{fig:cbt_nMostFreq_hist}
        \end{subfigure}
        \caption{
      Subfigure (a) shows the model accuracy when the correct answer is among $n$ most frequent named entities for $n\in [1,10]$. Subfigure (b) shows the number of test examples for which the correct answer was the $n$--th most frequent entity.}
        \label{fig:cbt_nMostFreq}
\end{figure}

\end{appendices}

\end{document}